\documentclass{article}

\usepackage[final]{corl_wcbm_2024} 

\usepackage[textsize=footnotesize]{todonotes}
\usepackage{amsmath}
\usepackage{amssymb}
\usepackage{mathtools}
\usepackage{amsthm}
\usepackage{bm}
\usepackage{glossaries}
\usepackage{graphicx}
\usepackage{multirow}
\usepackage{booktabs}
\usepackage{amsmath}
\usepackage{float}
\usepackage{wrapfig}

\newacronym{rl}{RL}{Reinforcement Learning}
\newacronym{drl}{DRL}{Deep Reinforcement Learning}
\newacronym{il}{IL}{Imitation Learning}
\newacronym{irl}{IRL}{Inverse Reinforcement Learning}
\newacronym{maxent}{MaxEnt}{Maximum Entropy}
\newacronym[plural=DPs, longplural=Diffusion Policies]{dp}{DP}{Diffusion Policy}
\newacronym{fm}{FM}{Foundation Model}
\newacronym{dr}{DR}{Domain Randomization}
\newacronym{sr}{SR}{Success Rate}

\newacronym{avi}{AVI}{Approximate Value-Iteration}
\newacronym{api}{API}{Approximate Policy-Iteration}
\newacronym[plural=MDPs, longplural=Markov Decision Processes]{mdp}{MDP}{Markov Decision Process}
\newacronym{cmdp}{CMDP}{Constrained Markov Decision Processes}
\newacronym{kl}{KL}{Kullback-Leibler Divergence}
\newacronym{gae}{GAE}{Generalized Advantage Estimation}
\newacronym{trpo}{TRPO}{Trust Region Policy Optimization}
\newacronym{ppo}{PPO}{Proxmimal Policy Optimization}
\newacronym{mpo}{MPO}{Maximum a Posteriori Policy Optimisation}
\newacronym{gans}{GANs}{Generative Adversarial Networks}
\newacronym{ail}{AIL}{Adversarial Imitation Learning}
\newacronym{gail}{GAIL}{Generative Adversarial Imitation Learning}
\newacronym{lsgail}{LS-GAIL}{Least-Squares Generative Adversarial Imitation Learning algorithms}
\newacronym{sqil}{SQIL}{Soft Q-Imitation Learning}
\newacronym{iq}{IQ-Learn}{Inverse soft Q-Learning}
\newacronym{sac}{SAC}{Soft-Actor Critic}
\newacronym{lsiq}{LS-IQ}{Least Squares Inverse Q-Learning}
\newacronym{lsgans}{LSGANs}{Least Squares Generative Adversarial Networks}
\newacronym{idm}{IDM}{Inverse-Dynamics Model}
\newacronym{deprl}{DEP-RL}{differential extrinsic plasticity}
\newacronym{dep}{DEP}{differential extrinsic plasticity}
\newacronym[plural=AMPs, longplural=Adversarial Motion Priors]{amp}{AMP}{Adversarial Motion Prior}

\newacronym{pd}{PD}{Proportional-Derivative}

\title{The Role of Domain Randomization in Training Diffusion Policies for Whole-Body Humanoid Control}

\author{%
Oleg Kaidanov$^{1,2}$, Firas Al-Hafez$^{1}$, Yusuf Süvari$^{1}$, Boris Belousov$^{2}$, Jan Peters$^{1,2,3}$ \\ \\
$^{1}$TU Darmstadt \quad ${^2}$German Research Center for AI (DFKI) \quad ${^3}$Hessian.AI \\
\texttt{oleg.kaidanov@dfki.de}
}

\begin{document}
\maketitle


\begin{abstract}
    Humanoids have the potential to be the ideal embodiment in environments designed for humans.
   Thanks to the structural similarity to the human body, they benefit from rich sources of demonstration data, e.g., collected 
   via teleoperation, motion capture, or even using videos of humans performing tasks.
    However, distilling a policy from demonstrations
    is still a challenging problem.
    While \glspl{dp} have shown impressive results in robotic manipulation, their applicability to locomotion and humanoid control remains underexplored.
    In this paper, we investigate how \emph{dataset diversity and size} affect the performance of~\glspl{dp} for humanoid whole-body control.
    In a simulated IsaacGym environment, we generate synthetic demonstrations by training \gls{amp} agents under various \gls{dr} conditions, and we compare \glspl{dp} fitted to datasets of different size and diversity. Our findings show that, although \glspl{dp} can achieve stable walking behavior, successful training of locomotion policies requires significantly larger and more diverse datasets compared to manipulation tasks, even in simple scenarios. Videos can be found at \url{https://sites.google.com/view/dps-for-humanoid-control}.
\end{abstract}

\keywords{Humanoid Control, Locomotion, Diffusion Policies} 

\glsresetall    


\section{Introduction}

Humanoid robots hold great promise as ideal embodiments for human-centered environments due to their structural resemblance to the human body, which enables them to leverage rich datasets like motion capture for learning control policies.
As more companies begin to develop, produce, and commercialize humanoid robots, there is a growing demand for robust and general whole-body motion policies. While diffusion models, in particular~\glspl{dp}, have recently achieved significant success in robot arm manipulation~\cite{chi2023diffusion,pearce2023imitating}, research on whole-body control --- encompassing simultaneous locomotion and arm movements --- remains relatively limited~\cite{ze2024generalizable,he2024omnih2o}. Most existing approaches that use diffusion models for robot control rely on small real-world datasets, which are often sparse and difficult to collect~\cite{reuss2023goal,ze20243d,ke20243d}.
In contrast, training~\gls{rl} policies for locomotion and whole-body control in randomized simulators has demonstrated substantial robustness and success~\cite{rudin2022learning,radosavovic2024real}.
Recent work, such as DiffuseLoco~\cite{huang2024diffuseloco}, has shown that diffusion models can integrate multiple source policies into a unified model for robot control.
However, while the focus has largely been on the development of the diffusion framework, the impact of the source dataset used for training remains underexplored.

To address the insufficient understanding of the role of dataset characteristics on training \glspl{dp}, we investigate different types of \gls{dr} --- such as perturbations, dynamic variations, and terrain changes --- as well as varying dataset sizes.
We collect datasets with distinct randomization strategies and sizes to train separate \glspl{dp}. Our findings reveal that \gls{dr} is crucial for training successful \glspl{dp}: even large datasets without sufficient randomization struggle to generalize to non-randomized environments.

\textbf{Contributions.} Our contributions are twofold. First, we present the first ablation study on the impact of \gls{dr} in dataset generation for training \glspl{dp} in humanoid control. This includes not only commonly used \gls{dr} techniques but also the introduction of a novel approach. Second, we analyze the effect of dataset size on training, exploring how varying amounts of data interact with different randomization techniques. Notably, while only a few trajectories may suffice for manipulation tasks, training \glspl{dp} for whole-body control demands substantially more data to achieve robust performance.

\section{Related Work}
\label{sec:related_work}
\textbf{Imitation Learning for Humanoid Control.}
One prominent approach in whole-body control is \gls{il} from motion capture data.
DeepMimic \cite{peng2018deepmimic} introduced an RL framework where physics-based characters learn skills by mimicking reference motions from a set of motion clips.
By combining imitation rewards with task-specific objectives, DeepMimic allows characters to both mimic reference motions and achieve specific goals. 
Most imitation learning approaches are based on \gls{gail} \cite{Ho2016}, in which a discriminator is learned and used as a reward signal to an \gls{rl} policy. Various adaptations have been proposed \cite{fu2018, Peng2021AMP, garg2021, alhafez2023}, with \gls{amp}~\cite{Peng2021AMP} achieving notable success. \gls{amp} leverages the discriminator's learned reward as a style reward, supplemented with additional handcrafted rewards for task-specific objectives. This approach encourages the agent to replicate the style of the dataset while effectively accomplishing the given tasks.
\citet{escontrela2022adversarial} demonstrated that \gls{amp} can produce natural locomotion strategies for quadrupedal robots using only a few seconds of motion capture data from a German Shepherd.
Extending these ideas to humanoid robots, however, presents additional challenges due to their higher degrees of freedom and balance requirements.
\citet{he2024learning} proposed Human-to-Humanoid (H2O), a framework for real-time whole-body teleoperation of humanoid robots using an RGB camera.
They introduced a ``sim-to-data'' process to filter and select feasible motions from a large human motion dataset. While their approach scales to a large number of motions, the dataset still consists of retargeted human motions without added variability. Our work builds upon this foundation by utilizing diffusion policy and versatile \gls{dr}. 

\textbf{Diffusion Policies in Robot Learning.}
Recent advancements in robotics have seen the integration of diffusion models into policy learning for legged locomotion tasks. Diffusion models, known for their ability to capture complex, multimodal distributions \cite{kang2023efficient}, have been effectively utilized to model the stochasticity and adaptability required for locomotion in varied environments.
Kang et al. \cite{kang2023efficient} propose Efficient Diffusion Policies (EDP) for offline \gls{rl}, aiming to overcome the computational inefficiency of previous diffusion-based approaches like Diffusion-QL. While their focus is on improving training efficiency and compatibility with various offline RL algorithms, the data collection in their experiments relies on existing offline datasets from benchmarks like D4RL~\cite{fu2020d4rl}, without specific emphasis on \gls{dr} or data diversity.

Ren et al. \cite{ren2024diffusion} introduce Diffusion Policy Policy Optimization (DPPO), a framework for fine-tuning diffusion-based policies using policy gradient methods. Their approach demonstrates improved performance and training stability over other \gls{rl} methods for diffusion policies. While their work focuses on fine-tuning pre-trained policies and does not delve deeply into the data collection process, they use \gls{dr} during sim-to-real transfer. By adding noise to observations and actions, they simulate imperfect conditions, thereby improving the robustness of their policies in real-world deployments. However, the application of \gls{dr}, in their approach compared to ours, is not extensively explored.

BiRoDiff introduced by Mothish et al. \cite{mothish2024birodiff} presents a real-time controller based on diffusion models for bipedal robots. They collect their source dataset by deploying a deep reinforcement learning policy trained via \gls{ppo} on their custom-made bipedal robot, Stoch BiRo. The dataset comprises observation-action pairs collected from walking on flat ground and slopes with specific inclinations. Their framework emphasizes generalization to unseen terrains, demonstrating the capability of diffusion policies to handle multiple walking behaviors with different velocities on various terrains. However, their approach primarily focuses on the generalization aspect without delving deeply into the challenges posed by dynamic whole-body control.

Huang et al. proposed DiffuseLoco \cite{huang2024diffuseloco}, a framework that leverages diffusion models to learn multi-skill locomotion policies from offline datasets. DiffuseLoco showcases the potential of diffusion policies in capturing diverse locomotion skills, including agile maneuvers like hopping and bipedal walking. To collect the source dataset, they generate data from multiple single-skill control policies obtained through various methods, including reinforcement learning and central pattern generators. They focus on offline learning, which allows them to scalably incorporate diverse skills. While their work highlights the scalability and versatility of diffusion models in robotics, it also addresses the importance of \gls{dr} in handling dynamic tasks involving whole-body motions.
    
While these works collectively advance the field of diffusion-based policies for locomotion, there is a gap in addressing the challenges posed by dynamic whole-body control tasks. \gls{dr} has been shown to be crucial in bridging the simulation-to-reality gap, particularly for tasks involving high dynamics and complex interactions \cite{tobin2017domain, james2022glide}.
Our approach builds upon the strengths of diffusion-based policies and explicitly integrates \gls{dr} to handle the complexities of dynamic whole-body control tasks. By introducing variability during training, our method enhances the robustness and adaptability of the learned policies, ensuring better performance in real-world scenarios with diverse and unpredictable conditions.
	
\section{Framework for Offline Dataset Generation}
\label{sec:dom_rand}
To generate datasets for training Diffusion Policies, we first train robust~\gls{rl} policies using~\gls{amp}~\cite{peng2021}.
This method integrates goal-conditioned \gls{rl} with \gls{il}. 
In our case, the goal is a velocity command.
For the imitation component, we leverage $10$ motion capture sequences from the AMASS dataset~\cite{mahmood2019amass}, which include walking in various directions (forward, backward, sideways), in-place rotations, walking in circles, and ``8-figure'' walking.

The policy is trained under extensive Domain Randomization (cf. Table~\ref{tab:domain_randomization_params}), incorporating several widely adopted regularization rewards (cf. Table~\ref{tab:reg_rewards}) to enhance the stability and smoothness of the resulting motions~\cite{rudin2022learning}.
%
Having trained the \gls{rl} policy, we utilize it to collect a total of $24$ distinct datasets ($3$ dataset sizes across $8$ environment setups), which are used to train diffusion models both individually and collectively.
The specific environment randomization configurations are detailed in Sec.~\ref{sec:rand_setups}.
Once the diffusion models are trained, we assess their performance across two distinct evaluation setups: i) without \gls{dr} on a flat surface, and ii) with dynamics randomization on a complex, uneven terrain.


\subsection{Reinforcement Learning Environment}
We model the environment as a \gls{mdp}, defined as a tuple 
$(\mathcal{S}, \mathcal{A}, \mathcal{T}, \mathcal{R}, p_0, \gamma)$. $\mathcal{S}$ denotes the state space, $\mathcal{A}$ represents the action space, $\mathcal{T}$ describes the transition function, $\mathcal{R}$ indicates the reward function, $p_0$ is the distribution of the initial state, and~$\gamma$ is the discount factor. The objective of \gls{rl} is to determine the optimal parameters $\theta$ of a parameterized policy $\pi_{\theta}: \mathcal{S} \to \mathcal{A}$ that maximizes the expected discounted return:
$
J(\theta) = \mathbb{E}_{\pi_{\theta}} \left[ \sum_{t=0}^{T-1} \gamma^t r_t \right]
$.

We use the \gls{amp} algorithm in a simulated Unitree H1 humanoid robot environment. Actions are represented by a $19$-dimensional vector $\mathbf{a}_t \in \mathcal{A} = \mathbb{R}^{19}$, 
which indicates the desired positional changes for each actuated joint, later processed by a 
\gls{pd} controller. Similar to prior work \cite{huang2024diffuseloco}, we give the critic privileged information such as base linear and angular velocities, while the policy receives a reduced observation vector without velocities. The privileged observations represented as $\mathbf{o}_t \in \mathbb{R}^{69}$, include the current linear and angular velocities of the 
humanoid, the orientation of the gravity vector relative to the robot’s base frame, joint positions 
and velocities, target command, and previous actions. A command vector $\mathbf{c}_t$ is used to define the target velocities along the x-, y-, and yaw-axis in the robot's base frame.

To encourage the robot to effectively track the desired command velocities $\hat{v}_{t,x}$, 
$\hat{v}_{t,y}$, and the global yaw rate $\hat{\omega}_t$, we define a task reward function similar to \cite{escontrela2022adversarial}, that assesses how closely the robot's actual motion aligns with these targets. Specifically, we want to minimize the difference between the robot's current linear and angular velocities $v_{t,x}$, $v_{t,y}$, $\omega_t$ and the desired values. The reward function $r_{\text{g}}$ at each timestep is defined as
\begin{equation}
r_{\text{g}} = w_v \exp \left( -\|\hat{v}_{t,xy} - v_{t,xy} \| \right) + w_\omega \exp \left( -|\hat{\omega}_t - \omega_t| \right).
\end{equation}
The desired velocities are sampled uniformly from the ranges: 
$\hat{v}_{t,x} \in (-1, 1) \ \mathrm{m/s}$, 
$\hat{v}_{t,y} \in (-0.7, 0.7) \ \mathrm{m/s}$, 
$\hat{\omega}_t \in (-1.57, 1.57) \ \mathrm{rad/s}$. This task reward function is combined with regularization rewards defined in Table \ref{tab:reg_rewards} and a style reward function learned by a discriminator in \gls{amp}.


\begin{table*}[t]
\vspace{-1em}
    \centering
    \begin{minipage}{.48\linewidth}
        \centering
        \caption{\small Regularization Rewards}
        \label{tab:reg_rewards}
        \resizebox{\linewidth}{!}{ 
        \begin{tabular}{c c c}
            \hline
            \textbf{Name} & \textbf{Value} & \textbf{Weight} \\
            \hline
            DoF lower limit & $-\max(\theta - \theta_{\text{lim,low}}, 0)$ & -4.0 \\
            DoF upper limit & $\min(\theta - \theta_{\text{lim,up}}, 0)$ & -4.0 \\
            DoF velocities & $\|\dot{\theta}\|$ & -3.0e-5 \\
            DoF acceleration & $\|\ddot{\theta}\|$ & -1.0e-7 \\
            Non-flat base orientation & $\|\mathbf{g}^{xy}\|$ & -1.0 \\
            \hline
        \end{tabular}
        }
    \end{minipage}
    \hspace{2em} 
    \begin{minipage}{.35\linewidth}
    \centering
    \caption{\small Dynamics Randomization}
    \label{tab:domain_randomization_params}
    \vspace{0.25em}
    \resizebox{\linewidth}{!}{ 
    \begin{tabular}{c c}
        \hline
        \textbf{Parameter} & \textbf{Range} \\
        \hline
        Body friction & [0.7, 1.3] \\
        Added base mass [kg] & [-2.0, 2.0] \\
        Link mass multiplier & [0.8, 1.2] \\
        PD gains multiplier & [0.8, 1.2] \\
        COM displacement [m] & [-0.15, 0.15] \\
        External perturbation [m/s] & [0, 0.6] \\
        Joint friction coeff. & [0.01, 1.15] \\
        Joint damping coeff. & [0.3, 1.5] \\
        \hline
    \end{tabular}
    }
    \end{minipage}%
\end{table*}

\subsection{Environment Randomizations During Data Collection}
\label{sec:rand_setups}

\gls{dr} is used in \gls{rl} to enhance an agent's robustness by training it across a distribution of similar environments, rather than just a single environment. This approach has proven crucial for addressing the sim-to-real gap in \gls{rl}, enabling zero-shot transfer of policies from simulation to the real world. In this work, we aim to apply \gls{dr} to randomize the source dataset used for training the \gls{dp}. To achieve this, we introduce various randomization clusters below, including both commonly used techniques and newly proposed ones. While we apply all \gls{dr} during the training of the \gls{amp} policy, we implement each one separately during the collection of the source dataset used to train the \gls{dp}. This approach enables us to identify the effects and importance of each randomization method.

\textbf{Dynamics randomization.} During dynamics randomization, we randomize the simulator's parameters, as initially proposed by \cite{peng2018DR}. A detailed overview of randomized parameters of the simulator is given in Table~\ref{tab:domain_randomization_params}.

\textbf{Perturbation randomization.} We apply force perturbations of random amplitude to the humanoid's torso, capping the maximum amplitude at $0.6$ m/s. These perturbations are introduced at 3-second intervals. The primary objective of perturbation randomization is to destabilize the robot's state, compelling it to adapt and learn effective recovery strategies from unstable conditions. By exposing the humanoid to various disturbances, we aim to enhance its resilience and improve its ability to maintain balance and perform tasks in dynamic environments. 


\textbf{Terrain randomization.} 
Instead of simulating a flat ground only, we randomize the terrain the humanoid is walking on. To do so, we utilize terrains supported in IsaacGym which include terrains with obstacles as well as bumpy surfaces.


\textbf{Initial state randomization.} Reference state initialization has proven important for imitating human movements \cite{peng2018deepmimic, Peng2021AMP}. Hence, we sample an initial state from the expert dataset at the beginning of each episode to initialize the simulation. To further diversify the source dataset, we sample random joint positions and velocities in a limited range at a $50\%$ chance.   


\textbf{Humanoid scale randomization.} 
\citet{alhafez2023_loco, alhafez2024} recently proposed an idea of using multiple body scales during training.
Here we aim to investigate how data collected from humanoids of different scales impacts the performance of the Diffusion Policy.
We initially trained three separate RL policies, each for a different scale. However, we later discovered that our RL policy, originally trained with a single scale, remains robust to a range of variations in the humanoid scale. As a result, we opted to use this policy for data collection.
When scaling is applied, the mass and inertia of all links are scaled by factors of $k^3$ and $k^5$, respectively, where $k$ is the scaling factor. To account for these changes, we scale the PD gains of the motors and the torque limits by a factor of $k^4$ for each scale. During data collection, humanoids of scales 0.8, 1.0, and 1.2 are used. 




\section{Diffusion Models for Whole-Body Humanoid Control}
\label{sec:diffusion}
In a recent work, \citet{huang2024diffuseloco} utilized diffusion models for learning skill transitions from a set of separately trained \gls{rl} skill policies.
Similarly, we adopt an encoder-decoder architecture, incorporating two 2-layer MLP encoders and six Transformer decoder layers, each having an 8-head cross-attention layer. The MLP encoders embed the previous state-action transitions and goal information, which --- along with the diffusion timestep embedding --- serve as conditional inputs.
Ground-truth actions, after the addition of noise, are passed through the Transformer decoder layers.
In each layer, cross-attention weights are computed between the noisy actions and the conditional information. Finally, the model minimizes the mean square error loss between the predicted and sampled noise. As proposed by \citet{huang2024diffuseloco}, we employ receding horizon control, where predictions are made for $n$ future steps, but only the first predicted action is executed.

\begin{figure}[t]
    \centering
    \includegraphics[width=0.85\linewidth]{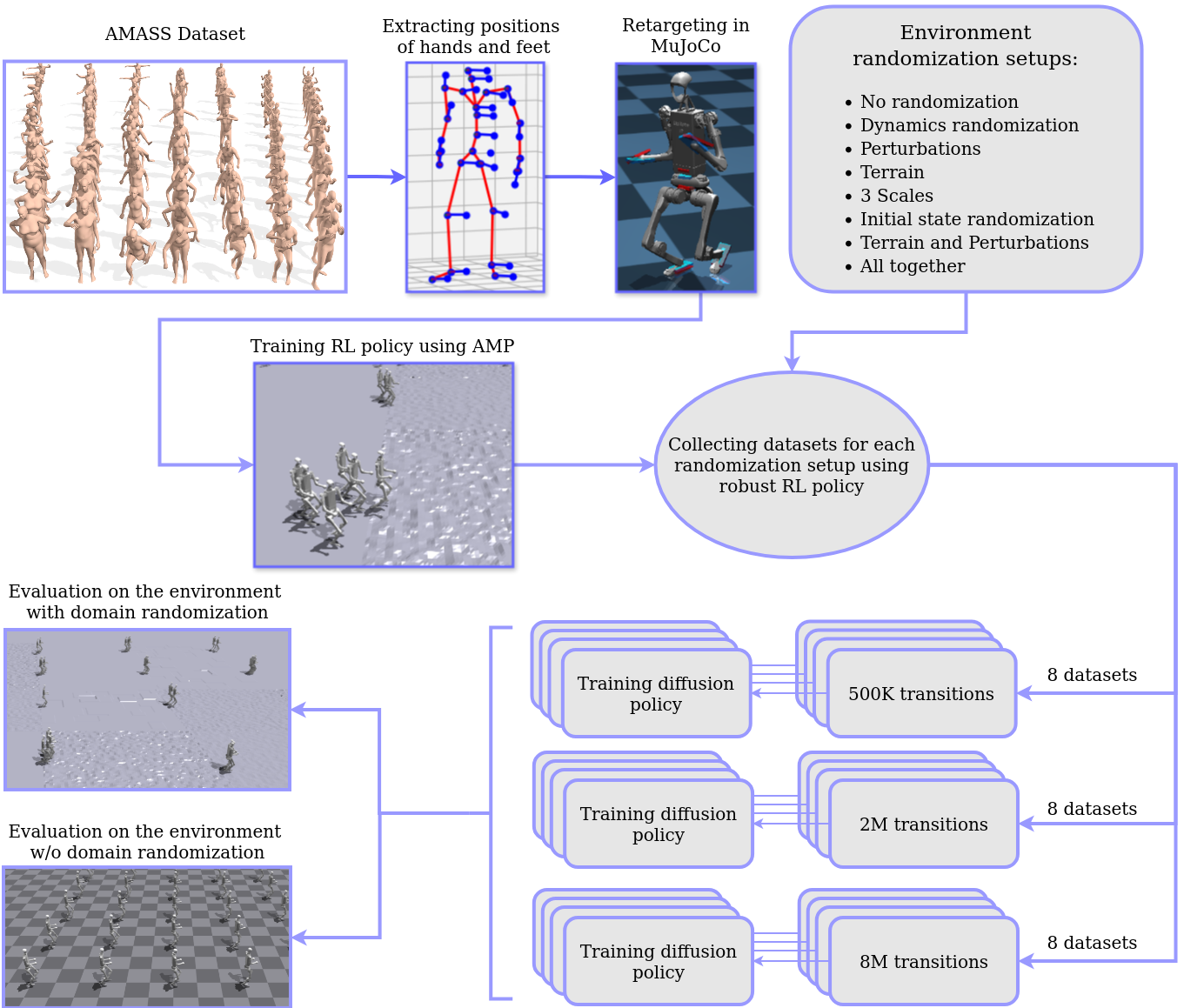} 
    \caption{Proposed method. First, a robust and stable \gls{rl} policy is trained using AMP under extensive Domain Randomization.
    This policy is then used for data collection, to subsequently train Diffusion Policies.
    We generate different datasets, each with different \gls{dr} applied during data collection, and we train \glspl{dp} on each dataset separately.
    Finally, performance of each~\gls{dp} is evaluated on two environments: with and without~\gls{dr}.}
    \label{fig:example_image}
    \vspace{-1em}
\end{figure}

Figure~\ref{fig:example_image} illustrates the overall setup. Motion sequences are collected from the AMASS dataset, from which the global positions and rotations of the hands and feet of a human subject are extracted. These keypoints are then processed using the MuJoCo simulator to solve the inverse kinematics problem, taking into account the humanoid morphology and potential collisions. This approach generates motion sequences consistent with humanoid structure, although not necessarily aligned with the physical dynamics of the real world or robot. These sequences, however, provide valuable style information, significantly reducing the need for manual reward shaping.
During training, this style information --- together with similarly structured observations from the simulator --- is provided to a discriminator within the \gls{amp} framework.
The discriminator evaluates how closely the humanoid motions in simulation resemble the preprocessed mocap data.
Once the RL policy is trained, it
serves as a data generator for subsequent training of diffusion-based policies.

\section{Results}
\label{sec:results}
\begin{figure}
    \centering
    \includegraphics[width=01.0\textwidth]{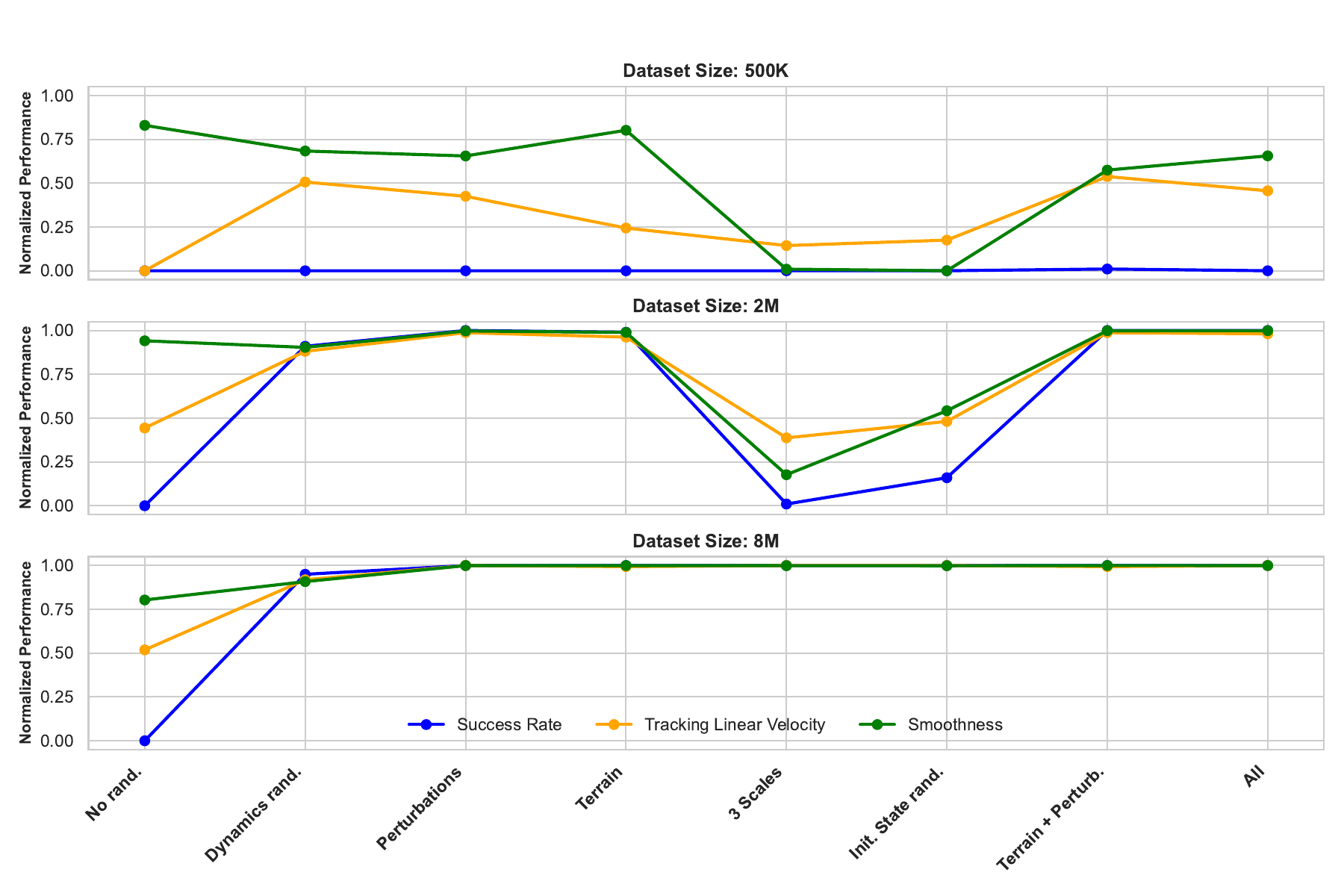}
    \resizebox{\textwidth}{!}{
    \begin{tabular}{c c c c c c c c c c}
        \hline
        \textbf{Dataset} & \textbf{Metric} & \textbf{No rand.} & \textbf{Dynamics rand.} & \textbf{Perturbations} & \textbf{Terrain} & \textbf{3 Scales} & \textbf{Init. State rand.} & \textbf{Terrain + Perturb.} & \textbf{All} \\
        \hline
        \multirow{3}{*}{\textbf{500K}} & Success Rate & 0.0 $\pm$ 0.0 & 0.0 $\pm$ 0.0 & 0.0 $\pm$ 0.0 & 0.0 $\pm$ 0.0 & 0.0 $\pm$ 0.0 & 0.0 $\pm$ 0.0 & 0.01 $\pm$ 0.0 & 0.0 $\pm$ 0.0 \\ 
        & Tracking lin. vel. & 1.73 $\pm$ 0.09 & 0.92 $\pm$ 0.11 & 1.05 $\pm$ 0.13 & 1.34 $\pm$ 0.06 & 1.5 $\pm$ 0.18 & 1.45 $\pm$ 0.12 & 0.87 $\pm$ 0.19 & 1.0 $\pm$ 0.18 \\ 
        & Smoothness & 49.36 $\pm$ 4.06 & 89.31 $\pm$ 13.48 & 97.02 $\pm$ 15.08 & 57.0 $\pm$ 4.65 & 273.25 $\pm$ 81.04 & 275.73 $\pm$ 33.22 & 119.02 $\pm$ 37.88 & 96.86 $\pm$ 27.21 \\
        \hline
        \multirow{3}{*}{\textbf{2M}} & Success Rate & 0.0 $\pm$ 0.0 & 0.91 $\pm$ 0.0 & \boldsymbol{$1.0$} $\pm$ \boldsymbol{$0.0$} & 0.99 $\pm$ 0.0 & 0.01 $\pm$ 0.0 & 0.16 $\pm$ 0.01 & \boldsymbol{$1.0$} $\pm$ \boldsymbol{$0.0$} & \boldsymbol{$1.0$} $\pm$ \boldsymbol{$0.0$} \\ 
        & Tracking lin. vel. & 1.02 $\pm$ 0.11 & 0.32 $\pm$ 0.05 & 0.15 $\pm$ 0.01 & 0.19 $\pm$ 0.07 & 1.11 $\pm$ 0.15 & 0.96 $\pm$ 0.24 & 0.15 $\pm$ 0.01 & 0.16 $\pm$ 0.01 \\ 
        & Smoothness & 18.93 $\pm$ 2.56 & 29.17 $\pm$ 10.0 & 3.61 $\pm$ 0.27 & 5.84 $\pm$ 19.5 & 227.38 $\pm$ 63.88 & 127.82 $\pm$ 60.74 & 3.24 $\pm$ 0.33 & 3.13 $\pm$ 0.17 \\ 
        \hline
        \multirow{3}{*}{\textbf{8M}} & Success Rate & 0.0 $\pm$ 0.0 & 0.95 $\pm$ 0.0 & \boldsymbol{$1.0$} $\pm$ \boldsymbol{$0.0$} & \boldsymbol{$1.0$} $\pm$ \boldsymbol{$0.0$} & \boldsymbol{$1.0$} $\pm$ \boldsymbol{$0.0$} & \boldsymbol{$1.0$} $\pm$ \boldsymbol{$0.0$} & \boldsymbol{$1.0$} $\pm$ \boldsymbol{$0.0$} & \boldsymbol{$1.0$} $\pm$ \boldsymbol{$0.0$} \\ 
        & Tracking lin. vel. & 0.9 $\pm$ 0.16 & 0.26 $\pm$ 0.07 & \boldsymbol{$0.13$} $\pm$ \boldsymbol{$0.0$} & 0.14 $\pm$ 0.01 & \boldsymbol{$0.13$} $\pm$ \boldsymbol{$0.0$} & 0.13 $\pm$ 0.01 & 0.14 $\pm$ 0.0 & \boldsymbol{$0.13$} $\pm$ \boldsymbol{$0.0$} \\ 
        & Smoothness & 56.51 $\pm$ 28.58 & 27.94 $\pm$ 6.12 & 2.97 $\pm$ 0.12 & 2.99 $\pm$ 0.1 & 3.11 $\pm$ 0.1 & 3.22 $\pm$ 0.21 & \boldsymbol{$2.9$} $\pm$ \boldsymbol{$0.05$} & 2.92 $\pm$ 0.07 \\ 
        \hline
    \end{tabular}
    }
    \caption{Evaluation of Diffusion Policies in a non-randomized target environment. \textbf{Top}: A plot displaying the normalized performances of all configurations, with tracking performance and smoothness inverted for unified metrics (higher values indicate better performance). \textbf{Bottom}: A table presenting detailed results, including the success rate (higher is better), tracking performance (lower is better), and smoothness (lower is better).}
    \label{fig:res_non_rand}
\end{figure}
We gathered a total of $24$ datasets, which comprise $8$ distinct environment setups and $3$ dataset sizes: $500$K, $2$M, and $8$M transitions of observations and actions. Each diffusion policy is trained with $3$ different random seeds. To assess the performance of the trained \gls{dp} across various environments, we create two validation environments: one without \gls{dr} and another with \gls{dr}. Each evaluation lasts for $10$ seconds, which corresponds to $500$ simulation steps. The commanding velocity is set to $1$ m/s in the forward direction. We evaluate the performance using three metrics: success rate, tracking performance, and smoothness. The success rate is defined as the number of environments that do not terminate before the end of the episode. The tracking performance is measured by the Euclidean distance between the current and commanded velocities. Smoothness is the sum of squared $L^2$ norms between two consecutive actions.
Training times on an Nvidia V100-16GB GPU are as follows: approximately 1 hour and 25 minutes for a dataset of 500,000 samples, about 5 hours for a dataset of 2 million samples, and roughly 23 hours and 17 minutes for a dataset of 8 million samples.

\subsection{Evaluation on Non-Randomized Target Environment}
\label{subsec:non-rand-res}

First, we evaluate \glspl{dp} in a non-randomized target environment. All \glspl{dp} are trained on datasets generated with various randomizations during data collection. Figure~\ref{fig:res_non_rand} presents the results for this setting. As observed, the \gls{dp} does not achieve stable walking, even when \gls{dr} is applied with a dataset size of $500$K samples. Starting from $2$M samples, some configurations achieve robust walking, although \gls{dr} becomes crucial for success. It is evident that some randomizations are more significant than others: while the configurations with dynamics, perturbations, and terrain randomization achieve a success rate close to $1.0$, scale and initial state randomizations do not yield successful policies. Nevertheless, with a large dataset size of $8$M, most randomizations produce strong results, with the notable exception of dynamics randomization, which achieves a success rate of $95$\%.

\begin{wraptable}{r}{0.4\textwidth}
\vspace{-1.75em}
    \centering
    \caption{Source \gls{rl} policy performance in a fixed target environment ($100$ runs)}
    \label{tab:baseline_eval_without_randomization}
    \resizebox{0.4\textwidth}{!}{
    \begin{tabular}{c c c }
        \hline
        \multirow{3}{*}{\textbf{Baseline}} & Success Rate & 1.0\\ 
        & Tracking lin. vel. & 0.12 $\pm$ 0.0\\ 
        & Smoothness & 2.89 $\pm$ 0.02\\ 
        \hline
    \end{tabular}
    }
\end{wraptable}
For reference, we report the performance of the source \gls{rl} policy in the same fixed target environment presented in Figure~\ref{fig:res_non_rand}. 
Comparing Table~\ref{tab:baseline_eval_without_randomization} with the 8M+All setting from Figure ~\ref{fig:res_non_rand}, we see that our best Diffusion Policy can match the performance of the source \gls{rl} policy.

\begin{figure}
    \centering
    \includegraphics[width=1.0\textwidth]{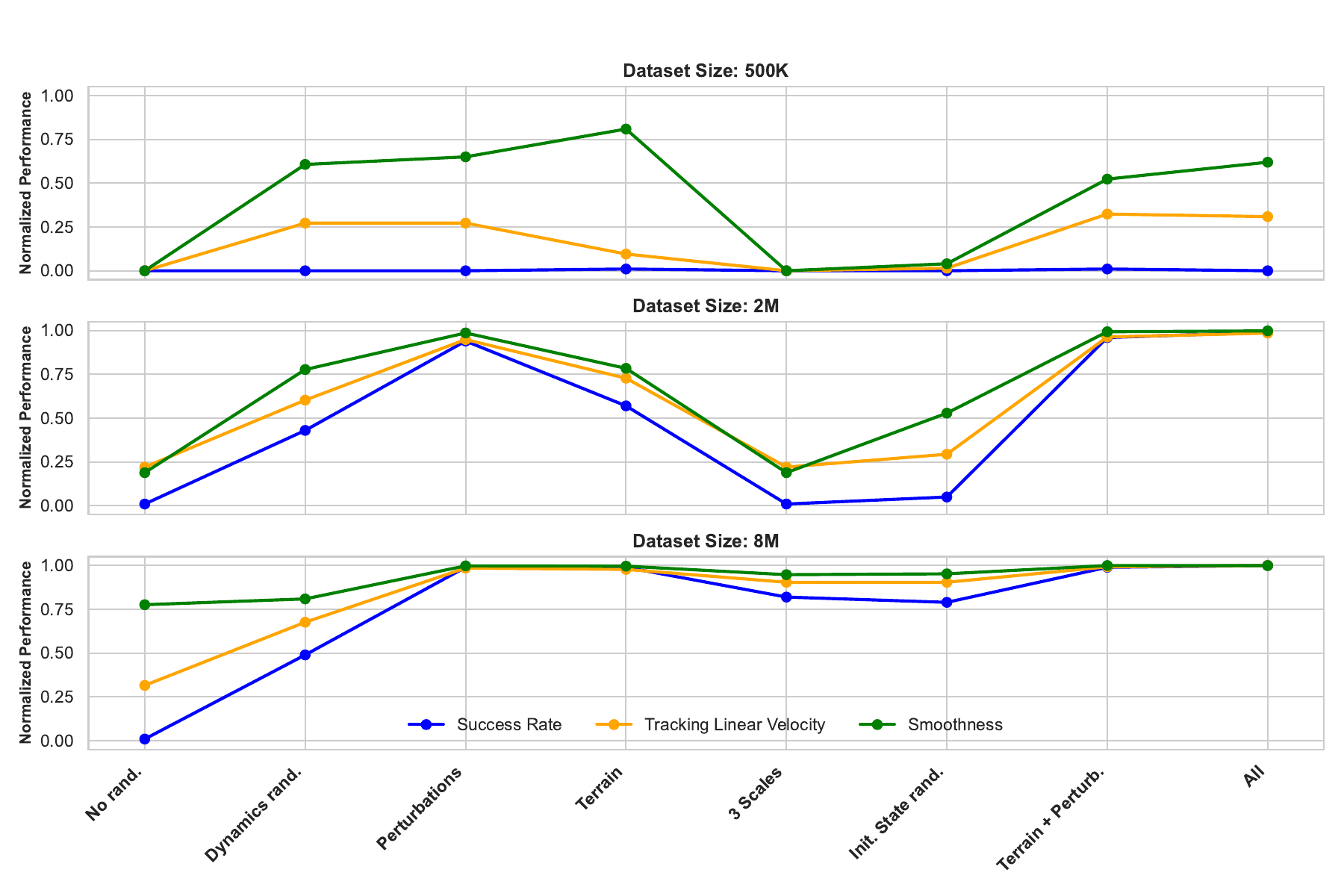}
     \resizebox{\textwidth}{!}{
    \begin{tabular}{c c c c c c c c c c}
        \hline
        \textbf{Dataset} & \textbf{Metric} & \textbf{No rand.} & \textbf{Dynamics rand.} & \textbf{Perturbations} & \textbf{Terrain} & \textbf{3 Scales} & \textbf{Init. State rand.} & \textbf{Terrain + Perturb.} & \textbf{All} \\
        \hline
        \multirow{3}{*}{\textbf{500K}} & Success Rate & 0.0 $\pm$ 0.0 & 0.0 $\pm$ 0.0 & 0.0 $\pm$ 0.0 & 0.01 $\pm$ 0.0 & 0.0 $\pm$ 0.0 & 0.0 $\pm$ 0.0 & 0.01 $\pm$ 0.0 & 0.0 $\pm$ 0.0 \\ 
        & Tracking lin. vel. & 1.5 $\pm$ 0.22 & 1.13 $\pm$ 0.27 & 1.13 $\pm$ 0.12 & 1.37 $\pm$ 0.12 & 1.5 $\pm$ 0.22 & 1.48 $\pm$ 0.16 & 1.06 $\pm$ 0.17 & 1.08 $\pm$ 0.16 \\ 
        & Smoothness & 277.23 $\pm$ 85.87 & 111.22 $\pm$ 21.34 & 99.39 $\pm$ 13.14 & 55.98 $\pm$ 10.53 & 277.23 $\pm$ 85.87 & 266.3 $\pm$ 37.24 & 134.12 $\pm$ 30.89 & 107.77 $\pm$ 27.75 \\
        \hline
        \multirow{3}{*}{\textbf{2M}} & Success Rate & 0.01 $\pm$ 0.0 & 0.43 $\pm$ 0.0 & 0.94 $\pm$ 0.0 & 0.57 $\pm$ 0.01 & 0.01 $\pm$ 0.0 & 0.05 $\pm$ 0.0 & 0.96 $\pm$ 0.0 & 0.99 $\pm$ 0.0 \\ 
        & Tracking lin. vel. & 1.2 $\pm$ 0.15 & 0.68 $\pm$ 0.47 & 0.21 $\pm$ 0.11 & 0.51 $\pm$ 0.38 & 1.2 $\pm$ 0.15 & 1.1 $\pm$ 0.2 & 0.19 $\pm$ 0.06 & 0.16 $\pm$ 0.03 \\ 
        & Smoothness & 225.54 $\pm$ 48.11 & 64.53 $\pm$ 42.46 & 7.41 $\pm$ 10.93 & 62.68 $\pm$ 78.58 & 225.54 $\pm$ 48.11 & 132.6 $\pm$ 51.38 & 5.54 $\pm$ 6.31 & 4.21 $\pm$ 1.11 \\
        \hline
        \multirow{3}{*}{\textbf{8M}} & Success Rate & 0.01 $\pm$ 0.0 & 0.49 $\pm$ 0.0 & 0.99 $\pm$ 0.0 & 0.99 $\pm$ 0.0 & 0.82 $\pm$ 0.0 & 0.79 $\pm$ 0.0 & 0.99 $\pm$ 0.0 & \boldsymbol{$1.0$} $\pm$ \boldsymbol{$0.0$} \\ 
        & Tracking lin. vel. & 1.07 $\pm$ 0.16 & 0.58 $\pm$ 0.44 & 0.16 $\pm$ 0.05 & 0.17 $\pm$ 0.05 & 0.27 $\pm$ 0.21 & 0.27 $\pm$ 0.19 & 0.15 $\pm$ 0.02 & \boldsymbol{$0.14$} $\pm$ \boldsymbol{$0.02$} \\ 
        & Smoothness & 64.77 $\pm$ 29.29 & 55.73 $\pm$ 34.79 & 4.15 $\pm$ 2.31 & 4.61 $\pm$ 7.17 & 17.89 $\pm$ 37.34 & 16.58 $\pm$ 26.01 & \boldsymbol{$3.62$} $\pm$ \boldsymbol{$0.9$} & 3.68 $\pm$ 0.69 \\
        \hline
    \end{tabular}
    }
    \caption{Evaluation of Diffusion Policies in a randomized target environment. Evaluation of Diffusion Policies in a non-randomized target environment. \textbf{Top}: A plot displaying the normalized performances of all configurations, with tracking performance and smoothness inverted for unified metrics (higher values indicate better performance). \textbf{Bottom}: A table presenting detailed results, including the success rate (higher is better), tracking performance (lower is better), and smoothness (lower is better).}
    \label{fig:res_rand}
\end{figure}

\subsection{Evaluation on Randomized Target Environment}
\label{subsec:rand-res}

Second, we evaluate \glspl{dp} in a randomized target environment. We generate this environment by applying milder domain randomization to provide a fair setup for all configurations. To do so, we applied only terrain and dynamics randomization, where we reduced the range of dynamics randomization. Detailed parameters are shown in  Table~\ref{tab:domain_randomization}. 

\begin{wraptable}{r}{0.35\textwidth}
\vspace{-0.75em}
    \centering
    \caption{\small \gls{dr} during evaluation}
    \label{tab:domain_randomization}
    \vspace{0.25em}
    \resizebox{0.35\textwidth}{!}{
    \begin{tabular}{c c}
        \hline
        \textbf{Parameter} & \textbf{Range} \\
        \hline
        Body friction & [0.8, 1.2] \\
        Added base mass [kg] & [-1., 1] \\
        Link mass multiplier & [0.9, 1.1] \\
        PD gains multiplier & [0.9, 1.1] \\
        COM displacement [m] & [-0.1, 0.1] \\
        Joint friction coeff. & [0.01, 0.5] \\
        Joint damping coeff. & [0.3, 1.] \\
        \hline
    \end{tabular}
    }
    \vspace{-1em}
\end{wraptable}
Once more, all \glspl{dp} are trained on datasets generated with various randomizations during data collection. When comparing these results to those from the non-randomized target environment, the impact of each randomization cluster becomes clearer. In the results using $500$K samples, no configuration achieves stable walking. However, starting from $2$M samples, perturbations yield strong results, while other configurations perform poorly. With a larger dataset size of $8$M samples, perturbation and terrain randomization demonstrate the strongest outcomes, whereas the other configurations perform less effectively, with dynamics randomization only achieving a success rate of $50$\%.
\begin{wraptable}{r}{0.4\textwidth}
\vspace{-1.8em}
    \centering
    \caption{Source \gls{rl} policy performance in $100$ randomized target environments}
    \label{tab:baseline_eval_with_randomization}
    \resizebox{0.4\textwidth}{!}{
    \begin{tabular}{c c c }
        \hline
        \multirow{3}{*}{\textbf{Baseline}} & Success Rate & 1.0\\ 
        & Tracking lin. vel. & 0.13$\pm$ 0.02\\ 
        & Smoothness & 3.45 $\pm$ 0.36\\ 
        \hline
    \end{tabular}
    }
    \vspace{-1em}
\end{wraptable}
%

Comparing \glspl{dp} performance to the source \gls{rl} policy, we again find that our best \gls{dp} is able to match the \gls{rl} policy performance, despite target environment randomizations  as can be seen in Table~\ref{tab:baseline_eval_with_randomization}.

\vspace{-0.5em}
\section{Discussion}
\vspace{-0.5em}
\label{sec:discussion}

Our results reveal two key findings. First, we find that \gls{dr} is essential across all dataset sizes, even when the \gls{dp} is evaluated in environments equivalent to those used for collecting the training dataset (see \ref{subsec:non-rand-res}). Second, we note that the dataset size also plays a significant role when applying \gls{dr}. This contrasts sharply with previous work in \gls{dp} for manipulation, where often only a few expert trajectories are sufficient to accomplish a task. This finding emphasizes the importance of both \gls{dr} and dataset size in complex and dynamic tasks, such as whole-body humanoid control.


We also found that not all randomizations are of equal importance. For some of the randomization setups, an increase of the dataset size from $2$M to $8$M leads to significantly better performance (cf. Figure~\ref{fig:res_non_rand} and Figure~\ref{fig:res_rand}, columns ``3 Scales'' and ``Init state rand.''). In other cases, however, the improvement in performance is marginal, and the overall performance is low (cf. Figure~\ref{fig:res_rand}, columns ``No rand.'' and Dynamics), highlighting the fact that the dataset size alone cannot compensate for the lack of diversity in the training data.

In the settings, where the data was collected with external perturbations and on different terrains (cf. Figure~\ref{fig:res_non_rand} and Figure~\ref{fig:res_rand}, columns Perturbations, Terrain+Perturbations and All), \glspl{dp} could achieve high success rate, tracking performance, and smoothness on the datasets of size $2$M. Increasing the dataset size to $8$M only slightly improved the results. 

Another interesting finding is that \gls{rl} policies trained under extensive \gls{dr} perform well on both smaller and larger body scales compared to the original setup. Despite the highly nonlinear changes in the robot dynamics parameters, such as the masses and inertias of the links, the policy maintains almost the same performance across different scales when the PD gains are scaled by a factor of {$k^4$}.

\vspace{-0.5em}
\section{Conclusion}
\vspace{-0.5em}
\label{sec:conclusion}
We evaluated the effects of dataset size and diversity on training of \glspl{dp} for whole-body humanoid control.
In particular, we investigated which randomization configurations have the largest impact on the training performance. We found that terrain and perturbation randomization are the most important configurations to ensure high data coverage and therefore good generalization performance of the trained \glspl{dp}. Additionally, we found that the dataset size plays a crucial role when learning highly dynamic tasks such as whole-body humanoid control.
%
We believe that our findings will provide helpful insights for further research efforts in whole-body humanoid control using \gls{dp}.

\vspace{-0.5em}
\acknowledgments{
\vspace{-0.5em}
This research was supported by the Research Clusters “The Adaptive Mind” and “Third Wave of AI,” funded by the Excellence Program of the Hessian Ministry of Higher Education, Science, Research and the Arts, and by the grant for establishing a DFKI lab at TU Darmstadt. We acknowledge computing time on the Lichtenberg II HPC at TU Darmstadt, funded by the BMBF and the State of Hesse. Partial support was also provided by the DFG under grant SE1042/41-1.
}


\bibliography{main}  

\begin{thebibliography}{29}
\providecommand{\natexlab}[1]{#1}
\providecommand{\url}[1]{\texttt{#1}}
\expandafter\ifx\csname urlstyle\endcsname\relax
  \providecommand{\doi}[1]{doi: #1}\else
  \providecommand{\doi}{doi: \begingroup \urlstyle{rm}\Url}\fi

\bibitem[Chi et~al.(2023)Chi, Xu, Feng, Cousineau, Du, Burchfiel, Tedrake, and Song]{chi2023diffusion}
C.~Chi, Z.~Xu, S.~Feng, E.~Cousineau, Y.~Du, B.~Burchfiel, R.~Tedrake, and S.~Song.
\newblock Diffusion policy: Visuomotor policy learning via action diffusion.
\newblock \emph{The International Journal of Robotics Research}, page 02783649241273668, 2023.

\bibitem[Pearce et~al.(2023)Pearce, Rashid, Kanervisto, Bignell, Sun, Georgescu, Macua, Tan, Momennejad, Hofmann, and Devlin]{pearce2023imitating}
T.~Pearce, T.~Rashid, A.~Kanervisto, D.~Bignell, M.~Sun, R.~Georgescu, S.~V. Macua, S.~Z. Tan, I.~Momennejad, K.~Hofmann, and S.~Devlin.
\newblock Imitating human behaviour with diffusion models.
\newblock In \emph{International Conference on Learning Representations (ICLR)}. OpenReview.net, 2023.

\bibitem[Ze et~al.(2024)Ze, Chen, Wang, Chen, He, Yuan, Peng, and Wu]{ze2024generalizable}
Y.~Ze, Z.~Chen, W.~Wang, T.~Chen, X.~He, Y.~Yuan, X.~B. Peng, and J.~Wu.
\newblock Generalizable humanoid manipulation with improved 3d diffusion policies.
\newblock \emph{arXiv preprint arXiv:2410.10803}, 2024.

\bibitem[He et~al.(2024)He, Luo, He, Xiao, Zhang, Zhang, Kitani, Liu, and Shi]{he2024omnih2o}
T.~He, Z.~Luo, X.~He, W.~Xiao, C.~Zhang, W.~Zhang, K.~Kitani, C.~Liu, and G.~Shi.
\newblock Omnih2o: Universal and dexterous human-to-humanoid whole-body teleoperation and learning.
\newblock \emph{arXiv preprint arXiv:2406.08858}, 2024.

\bibitem[Reuss et~al.(2023)Reuss, Li, Jia, and Lioutikov]{reuss2023goal}
M.~Reuss, M.~Li, X.~Jia, and R.~Lioutikov.
\newblock Goal-conditioned imitation learning using score-based diffusion policies.
\newblock \emph{arXiv preprint arXiv:2304.02532}, 2023.

\bibitem[Ze et~al.(2024)Ze, Zhang, Zhang, Hu, Wang, and Xu]{ze20243d}
Y.~Ze, G.~Zhang, K.~Zhang, C.~Hu, M.~Wang, and H.~Xu.
\newblock 3d diffusion policy.
\newblock \emph{arXiv preprint arXiv:2403.03954}, 2024.

\bibitem[Ke et~al.(2024)Ke, Gkanatsios, and Fragkiadaki]{ke20243d}
T.-W. Ke, N.~Gkanatsios, and K.~Fragkiadaki.
\newblock 3d diffuser actor: Policy diffusion with 3d scene representations.
\newblock In \emph{First Workshop on Vision-Language Models for Navigation and Manipulation at ICRA 2024}, 2024.

\bibitem[Rudin et~al.(2022)Rudin, Hoeller, Reist, and Hutter]{rudin2022learning}
N.~Rudin, D.~Hoeller, P.~Reist, and M.~Hutter.
\newblock Learning to walk in minutes using massively parallel deep reinforcement learning.
\newblock In \emph{Conference on Robot Learning}, pages 91--100. PMLR, 2022.

\bibitem[Radosavovic et~al.(2024)Radosavovic, Xiao, Zhang, Darrell, Malik, and Sreenath]{radosavovic2024real}
I.~Radosavovic, T.~Xiao, B.~Zhang, T.~Darrell, J.~Malik, and K.~Sreenath.
\newblock Real-world humanoid locomotion with reinforcement learning.
\newblock \emph{Science Robotics}, 9\penalty0 (89):\penalty0 eadi9579, 2024.

\bibitem[Huang et~al.(2024)Huang, Chi, Wang, Li, Peng, Shao, Nikolic, and Sreenath]{huang2024diffuseloco}
X.~Huang, Y.~Chi, R.~Wang, Z.~Li, X.~B. Peng, S.~Shao, B.~Nikolic, and K.~Sreenath.
\newblock Diffuseloco: Real-time legged locomotion control with diffusion from offline datasets.
\newblock \emph{arXiv preprint arXiv:2404.19264}, 2024.

\bibitem[Peng et~al.(2018)Peng, Abbeel, Levine, and van~de Panne]{peng2018deepmimic}
X.~B. Peng, P.~Abbeel, S.~Levine, and M.~van~de Panne.
\newblock Deepmimic: Example-guided deep reinforcement learning of physics-based character skills.
\newblock \emph{ACM Transactions on Graphics}, 37\penalty0 (4):\penalty0 143:1--143:18, 2018.
\newblock \doi{10.1145/3197517.3201311}.

\bibitem[Ho and Ermon(2016)]{Ho2016}
J.~Ho and S.~Ermon.
\newblock Generative adversarial imitation learning.
\newblock In \emph{Proceeding of the Thirtieth Conference on Neural Information Processing Systems}, Barcelona, Spain, Dec. 2016.

\bibitem[Fu et~al.(2018)Fu, Luo, and Levine]{fu2018}
J.~Fu, K.~Luo, and S.~Levine.
\newblock Learning robust rewards with adversarial inverse reinforcement learning.
\newblock In \emph{Proceeding of the International Conference on Learning Representations}, Vancouver, Canada, Apr. 2018.

\bibitem[Peng et~al.(2021)Peng, Ma, Abbeel, Levine, and Kanazawa]{Peng2021AMP}
X.~B. Peng, Z.~Ma, P.~Abbeel, S.~Levine, and A.~Kanazawa.
\newblock Amp: Adversarial motion priors for stylized physics-based character control.
\newblock \emph{ACM Transactions on Graphics}, 40\penalty0 (4):\penalty0 1--20, 2021.
\newblock \doi{10.1145/3450626.3459670}.

\bibitem[Garg et~al.(2021)Garg, Chakraborty, Cundy, Song, and Ermon]{garg2021}
D.~Garg, S.~Chakraborty, C.~Cundy, J.~Song, and S.~Ermon.
\newblock Iq-learn: Inverse soft-q learning for imitation.
\newblock In \emph{Proceeding of the Thirty-fifth Conference on Neural Information Processing Systems}, Sydney, Australia, Dec. 2021.

\bibitem[Al-Hafez et~al.(2023)Al-Hafez, Tateo, Arenz, Zhao, and Peters]{alhafez2023}
F.~Al-Hafez, D.~Tateo, O.~Arenz, G.~Zhao, and J.~Peters.
\newblock {LS-IQ}: Implicit reward regularization for inverse reinforcement learning.
\newblock In \emph{Proceeding of the International Conference on Learning Representations}, Kigali, Rwanda, May 2023.

\bibitem[Escontrela et~al.(2022)Escontrela, Peng, Yu, Zhang, Iscen, Goldberg, and Abbeel]{escontrela2022adversarial}
A.~Escontrela, X.~B. Peng, W.~Yu, T.~Zhang, A.~Iscen, K.~Goldberg, and P.~Abbeel.
\newblock Adversarial motion priors make good substitutes for complex reward functions.
\newblock \emph{arXiv preprint arXiv:2203.15103}, 2022.

\bibitem[He et~al.(2024)He, Luo, Xiao, Zhang, Kitani, Liu, and Shi]{he2024learning}
T.~He, Z.~Luo, W.~Xiao, C.~Zhang, K.~Kitani, C.~Liu, and G.~Shi.
\newblock Learning human-to-humanoid real-time whole-body teleoperation.
\newblock \emph{arXiv preprint arXiv:2403.07208}, 2024.

\bibitem[Kang et~al.(2023)Kang, Ma, Du, Pang, and Yan]{kang2023efficient}
B.~Kang, X.~Ma, C.~Du, T.~Pang, and S.~Yan.
\newblock Efficient diffusion policies for offline reinforcement learning.
\newblock In \emph{Advances in Neural Information Processing Systems}, 2023.

\bibitem[Fu et~al.(2020)Fu, Kumar, Nachum, Tucker, and Levine]{fu2020d4rl}
J.~Fu, A.~Kumar, O.~Nachum, G.~Tucker, and S.~Levine.
\newblock D4rl: Datasets for deep data-driven reinforcement learning.
\newblock \emph{arXiv preprint arXiv:2004.07219}, 2020.

\bibitem[Ren et~al.(2024)Ren, Lidard, Ankile, Simeonov, Agrawal, Majumdar, Burchfiel, Dai, and Simchowitz]{ren2024diffusion}
A.~Z. Ren, J.~Lidard, L.~L. Ankile, A.~Simeonov, P.~Agrawal, A.~Majumdar, B.~Burchfiel, H.~Dai, and M.~Simchowitz.
\newblock Diffusion policy policy optimization.
\newblock \emph{arXiv preprint arXiv:2409.00588}, 2024.

\bibitem[Mothish et~al.(2024)Mothish, Tayal, and Kolathaya]{mothish2024birodiff}
G.~V.~S. Mothish, M.~Tayal, and S.~Kolathaya.
\newblock Birodiff: Diffusion policies for bipedal robot locomotion on unseen terrains.
\newblock \emph{arXiv preprint arXiv:2407.05424}, 2024.

\bibitem[Tobin et~al.(2017)Tobin, Fong, Ray, Schneider, Zaremba, and Abbeel]{tobin2017domain}
J.~Tobin, R.~Fong, A.~Ray, J.~Schneider, W.~Zaremba, and P.~Abbeel.
\newblock Domain randomization for transferring deep neural networks from simulation to the real world.
\newblock In \emph{2017 IEEE/RSJ International Conference on Intelligent Robots and Systems (IROS)}, pages 23--30, 2017.

\bibitem[James et~al.(2022)James, Makoviychuk, Yu, Tan, and Fox]{james2022glide}
A.~M. James, A.~Makoviychuk, T.~W. Yu, J.~Tan, and D.~Fox.
\newblock Glide: Generalizable quadrupedal locomotion in diverse environments with a centroidal model.
\newblock \emph{Robotics: Science and Systems}, 2022.

\bibitem[Peng et~al.(2021)Peng, Ma, Abbeel, Levine, and Kanazawa]{peng2021}
X.~B. Peng, Z.~Ma, P.~Abbeel, S.~Levine, and A.~Kanazawa.
\newblock Amp: Adversarial motion priors for stylized physics-based character control.
\newblock \emph{ACM Transactions on Graphics}, 40:\penalty0 1 -- 20, 2021.

\bibitem[Mahmood et~al.(2019)Mahmood, Ghorbani, Troje, Pons-Moll, and Black]{mahmood2019amass}
N.~Mahmood, N.~Ghorbani, N.~F. Troje, G.~Pons-Moll, and M.~J. Black.
\newblock Amass: Archive of motion capture as surface shapes.
\newblock In \emph{Proceedings of the IEEE/CVF international conference on computer vision}, pages 5442--5451, 2019.

\bibitem[Peng et~al.(2018)Peng, Andrychowicz, Zaremba, and Abbeel]{peng2018DR}
X.~B. Peng, M.~Andrychowicz, W.~Zaremba, and P.~Abbeel.
\newblock Sim-to-real transfer of robotic control with dynamics randomization.
\newblock In \emph{Proceeding of the IEEE International Conference on Robotics and Automation (ICRA)}, Brisbane, Australia, May 2018.

\bibitem[Al-Hafez et~al.(2023)Al-Hafez, Zhao, Peters, and Tateo]{alhafez2023_loco}
F.~Al-Hafez, G.~Zhao, J.~Peters, and D.~Tateo.
\newblock {LocoMuJoCo}: A comprehensive imitation learning benchmark for locomotion.
\newblock In \emph{6th Robot Learning Workshop, NeurIPS}, New Orleans, Louisiana, United States, Dec. 2023.

\bibitem[Al-Hafez et~al.(2024)Al-Hafez, Zhao, Peters, and Tateo]{alhafez2024}
F.~Al-Hafez, G.~Zhao, J.~Peters, and D.~Tateo.
\newblock Time-efficient reinforcement learning with stochastic stateful policies.
\newblock In \emph{Proceeding of the International Conference on Learning Representations}, Vienna, Austria, May 2024.

\end{thebibliography}

\end{document}